\documentclass{bmvc2k}

\usepackage{times}
\usepackage{epsfig}
\usepackage{graphicx}
\usepackage{amsmath}
\usepackage{amssymb}
\usepackage{xcolor}
\usepackage{subfigure}
\usepackage{mathrsfs}
\usepackage{comment}
\usepackage{amsfonts}
\usepackage{pifont}
\usepackage{array}
\usepackage{multirow}
\usepackage{multicol}
\newcommand{\cmark}{\ding{51}}%
\newcommand{\xmark}{\ding{55}}%
\newcommand{\bz}{\mathbf{z}}
\newcommand*{\eg}{\emph{e.g.}\@\xspace}
\newcommand*{\ie}{\emph{i.e.}\@\xspace}

\title{Local and Global Point Cloud Reconstruction for 3D Hand Pose Estimation}
\addauthor{Ziwei Yu}{yuziwei@u.nus.edu}{1}
\addauthor{Linlin Yang\textsuperscript{1, }}{yangll@comp.nus.edu.sg}{2}
\addauthor{Shicheng Chen}{e0534721@u.nus.edu}{1}
\addauthor{Angela Yao}{ayao@comp.nus.edu.sg}{1}

\addinstitution{
National University of Singapore,\\
Singapore
}
\addinstitution{
University of Bonn,\\
Germany
}

\runninghead{Z. Yu ET AL.}{Local and Global Point Cloud Reconstruction}
\def\eg{\emph{e.g}\bmvaOneDot}

\def\etal{\emph{et al}\bmvaOneDot}
\begin{document}

\maketitle

\begin{abstract}
This paper addresses the 3D point cloud reconstruction and 3D pose estimation of the human hand from a single RGB image.
To that end, we present a novel pipeline for local and global point cloud reconstruction using a 3D hand template 
while learning a latent representation for pose estimation. To demonstrate our method, we introduce a new multi-view hand posture dataset to obtain complete 3D point clouds of the hand in the real world. Experiments on our newly proposed dataset and four public benchmarks demonstrate the model's strengths. Our method outperforms competitors in 3D pose estimation while reconstructing realistic-looking complete 3D hand point clouds.
\end{abstract}

\section{Introduction}
The 3D shape and pose of the human hand are critical for augmented and virtual reality applications. To accommodate this form of human-computer interaction, an entire discipline of computer vision is devoted to estimating 3D hand shape and pose. Achieving accurate estimates is extremely challenging due to the hand's high degrees of articulation and self-occlusion. Earlier approaches attempted to combine representations from various viewpoints~\cite{ge2016robust, ge20173d, ge2018point,ge2018hand}, or transform 2.5D depth maps to 3D representations such as voxels~\cite{moon2018v2v,malik2020handvoxnet}, point clouds~\cite{li2019point}, or meshes~\cite{wan2019dual}. Since 3D voxel models are computationally more expensive than mesh and point cloud models, the latter two are preferable for estimating 3D hand shape and pose.

Current RGB-based methods~\cite{boukhayma20193d,zhang2019end,kulon2020weakly} prefer to estimate hand shape by mapping visual features to the parameters of a parametric model \eg~MANO~\cite{romero2017embodied}. However, the MANO mesh inherently differs from the real hand surface, resulting in an unnaturally smoothed hand shape. Non-parametric mesh techniques~\cite{ge20193d,wan2019dual} can generate more realistic shapes and account for shape surface details but learning such models requires a substantial amount of annotated mesh data that are non-trivial to collect. 

We believe that 3D hand point clouds could serve as a non-parametric alternative to meshes. {Unlike meshes, point clouds are unordered and do not require predefined topological structure for surface.} Point clouds are easily obtained from various sources, \eg depth cameras, laser scanners, and other 3D representations. Moreover, a 3D hand point cloud's density is easily adjustable; depending on the resolution requirements, we can down- or up-sample the number of points from the surface of the hand.  The use of point clouds for 3D hand pose and shape estimation is limited~\cite{yang2019aligning,li2019point,ge2018point,ge2018hand}. The work most closely related to ours is~\cite{yang2019aligning}, which estimates a point cloud from different modalities, including RGB images. Their estimated point cloud, is only of the camera-facing surface. 
A complete point cloud would provide more complete geometric information, as shown previously in 3D body pose estimation~\cite{yao2019densebody,zhang2020object,alldieck2019tex2shape}.

This paper proposes a unique point cloud reconstruction technique for determining the full hand shape and pose from RGB images. We put forth a combined local and global representation for learning a detailed 3D latent representation for the hand. To reconstruct an accurate and high-resolution point cloud of the hand, we draw inspiration from point cloud architectures like FoldingNet~\cite{yang2018foldingnet} and AtlasNet~\cite{groueix1802atlasnet}.  However, our work is novel in that we design a new template initialization specifically for 3D hand recovery. Our template is flexible and enables us to pre-distribute the 3D points in a configuration more useful for reconstructing the 3D hand.
Additionally, we offer a semantic grouping strategy for reconstructing the local and global point clouds that correspond to the individual fingers.

Existing RGB-based 3D hand pose datasets do not have any corresponding (complete) 3D point cloud data. As such, we sample from the surface of the MANO mesh model to generate point clouds for existing datasets.  Additionally, we introduce a new multi-view RGB-D dataset and illustrate the usefulness of our methodology on real-world point clouds recovered from depth images to validate our methodology on real-world data.  

Our contributions are summarized as follows: 
\begin{itemize}
\item We propose a unique framework for 3D point cloud reconstruction of the hand with a customized 3D hand template.  
To our knowledge, our system is the first to reconstruct a complete 3D hand point cloud rather than just the camera-facing surface. 
\item  We propose an effective combined local and global point cloud reconstruction method which captures more detailing than a single global model. 
\item We introduce a multi-view RGB-D hand pose dataset with 3D joint annotations, fitted MANO parameters and depth-map based 3D point clouds.
\item Evaluation on four public benchmarks and our own newly proposed dataset verifies that our framework can outperform state-of-the-art approaches for 3D hand pose estimation while being able to reconstruct high-quality point clouds. 

\end{itemize}
 
\section{Related Works}
\textbf{Point Cloud Reconstruction} methods in deep learning primarily learn unordered representations by examining the intrinsic 3D structure. Tree-based models~\cite{klokov2017escape, zeng20183dcontextnet,gadelha2018multiresolution} represent point clouds through $k$-d trees. Other works propose innovative network designs, including PointNet \cite{qi2017pointnet}, PointCNN~\cite{li2018pointcnn} and RNN-based models~\cite{ye20183d}. To date, most of these approaches~\cite{yang2018foldingnet, qi2017pointnet, klokov2017escape,ye20183d} concentrate on point cloud reconstruction from 3D Lidar images. Our purpose, in comparison, is to recreate point clouds from RGB images.

\textbf{3D Hand pose estimation} often use depth maps or RGB pictures as input. The 2.5D-depth map is fed into deep neural networks~\cite{newell2016stacked} to obtain heatmaps which are then lifted to a 3D-hand joint location~\cite{sinha2016deephand,moon2018v2v,ge2016robust, ge20173d, ge2018point,ge2018hand,xiong2019a2j}. Recent papers attempt to estimate the hand pose of an RGB image along with additional modalities (\eg., depth or mask) as weak labels~\cite{cai2018weakly,yang2019aligning} or as intermediate forms of supervision~\cite{zimmermann2017learning,iqbal2018hand,zhang2019end}. Others~\cite{ge20173d, ge2018point,li2019point} have tried to convert depth maps to obtain an incomplete point cloud of the camera-facing viewpoint to help predict 3D pose. To the best of our knowledge, we are the first to integrate point cloud reconstruction with 3D pose estimation from monocular RGB images.

\textbf{3D Hand shape estimation} typically in the form of meshes, can be more challenging than pose estimation because it needs to simultaneously predict the mesh topology and vertex locations. Most approaches~\cite{joseph2016fits,khamis2015learning,mueller2019real,zhang2019end,kulon2020weakly} reconstruct the mesh by leveraging the parametric MANO model~\cite{romero2017embodied}. Using MANO allows these works to directly regress shape (and pose) parameters, which sit in a more tractable and lower-dimensional space.  For RGB inputs, the standard approach~\cite{zhang2019end, kulon2020weakly} is to firstly estimate 2D joint locations and then iteratively regress the MANO pose and shape parameters. Non-parametric methods leverage either a fully convolutional network~\cite{wan2019dual} or a graph convolutional network~\cite{ge20193d} to directly regress mesh vertices. 
Unlike these above approaches, we aim to recover complete point clouds, which we believe to be a more flexible 3D representation than a mesh. 
\section{Methodology}
\subsection{RGB Encoder and 3D Pose Decoder} \label{Pose Estimation} 
\begin{figure}[!htb]
\centering
\includegraphics[scale=0.27]{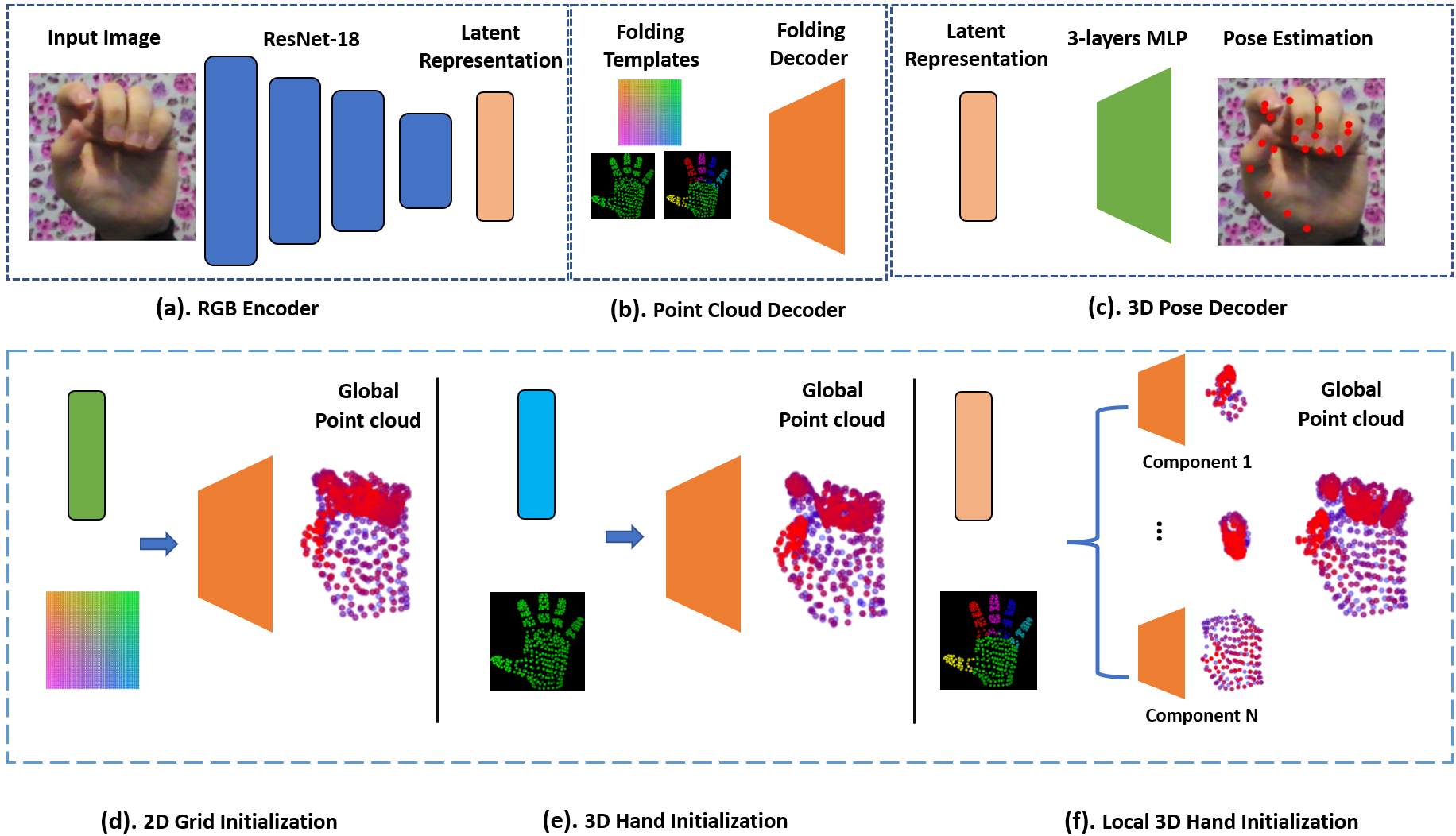}
\caption{\textbf{Overview of our point cloud reconstruction and pose estimation pipeline}.Our proposed has an RGB image encoder(a), a point cloud decoder (b) and a 3D pose decoder (c).  We experiment with three template initializations in (d) to (f) and observe that the most effective is to use local 3D initializations that represent the various semantic components of the hand (f).}
\label{fig:ourpipeline}
\end{figure}

Our framework has an RGB image encoder, a point cloud decoder, and a 3D pose decoder (see Fig.~\ref{fig:ourpipeline}). The encoder converts the image into a latent representation.
Our core contribution is in learning this latent representation such that it is rich and expressive for accurate 3D pose estimation (Sec. \ref{Pose Estimation}) and complete 3D point cloud reconstruction (Sec. \ref{Point Cloud Reconstruction}).  

The image encoder encodes a single $256\! \times\!256$ RGB image $\textbf{x}$  into a latent representation $\textbf{z $\in R^{512}$}$. The pose decoder converts $\textbf{z}$ into the hand pose $\textbf{J} \in \mathbb{R}^{3 \times 21}$,~\ie the 3D coordinates of 21 hand joints. We use the same backbone models as other encoder-decoder frameworks~\cite{kulon2020weakly,yang2019disentangling, yang2019aligning}.  For the encoder, we fine-tune a ResNet-18 backbone and use the final layer output fed into a fully connected layer as the latent representation $\textbf{z}$. For the 3D pose decoder, we use a three-layer fully connected multi-layer perceptron (MLP) with 128 hidden units per layer. To train the pose decoder, we use an L2 loss,~\ie, $\mathcal{L}_{Pose} = ||\textbf{J}_{pred}- \textbf{J}_{gt}||_2$, 
where the $\textbf{J}_{gt}$ and ${\textbf{J}_{pred}}$ denote the ground truth and estimated hand pose, respectively.
\subsection{3D Point Cloud Reconstruction} \label{Point Cloud Reconstruction}
To reconstruct a 3D hand point cloud $\hat{\mathcal{S}} \in \mathcal{R}^{N \times 3}$ from $\bz$, we follow the decoding architecture of FoldingNet~\cite{yang2018foldingnet}. FoldingNet's decoder is a series of MLPs that deform or ``fold'' a template set of points into their final 3D positions by conditioning on the encoded latent representation. Assuming that we are given an RGB image and an accompanying ground truth point cloud $\mathcal{S}$, the encoder-decoder pair can be learned via the Chamfer distance (\textbf{CD}) and Earth Mover's distance (\textbf{EMD}).  Both distances are computed between $\mathcal{S}$ and estimated point cloud set $\hat{\mathcal{S}}$ ($|\mathcal{S}|$ =  $|\hat{\mathcal{S}}|$ ). If $s \in R^{3 \times 1}$ and $\hat{s} \in R^{3 \times 1 }$ are ground truth and predicted 3D points respectively, then the Chamfer distance $d_{\text{CD}}$ is defined as
\begin{equation}
    d_{CD}(\mathcal{S}, \hat{\mathcal{S}})\!=\!\frac{1}{|S|} \sum_{s \in S} \min_{\hat{s} \in \hat{S}} ||s\!-\!\hat{s}||_2^2  + \frac{1}{|\hat{S}|} \sum_{\hat{s} \in \hat{S}} \min_{s \in S} ||s\!-\!\hat{s}||_2^2,
\label{equ:cd}
\vspace{-0.1cm}
\end{equation}
\noindent where the first term is the average distance of all predicted points to the closest ground truth point and the second term is the average distance of all ground truth points to the closest predicted point. The Earth-Mover's distance $d_{\text{EMD}}$ factors in the point-to-point assignment problem. Let $\phi: \hat{S} \rightarrow S$ be a bijection,~\ie for all $s \in S$, there is a uniquely matched point $s \in \hat{S}$. The optimal bijection is unique and invariant over the above point sets.
\begin{equation}
    d_{\text{EMD}}(\hat{\mathcal{S}},\mathcal{S}) = \min_{\phi: \hat{S} \rightarrow S} \sum_{s \in \hat{S}}||s - \phi(s)||_2.
\label{equ:emd}
\end{equation}

\textbf{Initialization Templates:} The original FoldingNet initializes the template point set on a 2D lattice grid (see Fig.~\ref{fig:ourpipeline}(d)).  The decoder then ``folds'' these points into a 3D surface structure. Using a 2D lattice grid is well-suited for class- or 3D-shape-agnostic reconstruction since it makes no prior assumptions.  However, we posit that it is non-ideal and inefficient for 3D hand shape estimation. It is more direct to initialize the point set to follow some canonical 3D hand. Therefore, we propose a 3D hand template initialization (see Fig.~\ref{fig:ourpipeline}(e)).  By fixing the initial spatial distribution to follow a hand, we simplify the learning of the decoder as it reduces the extent of folding required. The comparison of different initialization can be found in Supplementary Sec.~3.

\textbf{Local Reconstruction:} We further modify the FoldingNet decoder to do reconstruction locally.
In particular, we are interested in high-fidelity reconstructions of the fingers because they contain much of the pose information. It is therefore intuitive to offer separate representations to each of the fingers. To that end, we assign a local decoder to the palm and each of the fingers (see Fig.~\ref{fig:ourpipeline}(f)) and apply the Chamfer and Earth Mover's distance to these local point cloud sets. We leverage the (ordered) MANO vertices to separate the point cloud into individual components and their associated ground truths (see Sec.~\ref{vertice: gt}). To ensure that the components fit together, we also apply the two distances globally across the complete point set to arrive at the loss in Eq.~\ref{eq:total_loss}.

Similar to other point cloud reconstruction methods~\cite{fan2017point,lang2020samplenet,mandikal20183d, yang2019aligning}, we use the above two distance to learn the point cloud reconstruction.
Specifically, we apply the distances in a global sense,~\ie $d^G$ to the point cloud set of the entire hand, as well as in a local manner,~\ie $d^L$ to a subset of points that correspond to the local components. The final loss on the point cloud $\mathcal{L}_{\text{pc}}$ is a sum of these distances, \ie
\begin{equation}~\label{eq:total_loss}
     \mathcal{L}_{\text{pc}} = \sum_{i} \left(d_{CD}^{L_i} + d_{EMD}^{L_i} \right) + d_{CD}^{G} + d_{EMD}^{G},
\end{equation}
\noindent where $i$ indexes the local components.  While we can introduce weighting hyperparameters to the terms in Eq.~\ref{eq:total_loss}, we keep them equally weighted out of simplicity and as what have been done in~\cite{yang2019aligning}.

\subsection{Generating Ground Truth Point Clouds} \label{vertice: gt}
Existing RGB-based hand pose benchmarks have ground truth 3D poses but no point cloud information. As an alternative, we leverage the MANO~\cite{romero2017embodied} model and sample from the fitted mesh surface to obtain a set of 3D points. More specifically, MANO parameterizes a triangular mesh $\mathcal{M} \in R^{N \times 3}$ with parameters $\{ \vec{\beta}, \vec{\theta} \}$, where $\vec{\beta} \in R^{10}$ signify the shape parameters and $\vec{\theta} \in R^{K \times 3}$ are pose parameters. Similar to \cite{kulon2020weakly}, we fit the MANO model $\textbf{J}_{\text{MANO}} \in R^{21 \times 3 }$ to the ground truth 3D poses $\textbf{J}_{\text{gt}} \in R^{21 \times 3 } $ by minimizing the following L2 objective: $\min_{{\vec{\beta}, \vec{\theta}}}||\textbf{J}_{\text{gt}} - \textbf{J}_{\text{MANO}}||_2$.
Based on the fitted mesh vertices, we upsample and then randomly downsample to obtain a set of 3D points distributed on the hand surface.

\begin{figure}[ht]
    \centering
    \subfigure[MANO-generated]{
     \includegraphics[width =  0.20\textwidth, height = 0.16\textwidth]{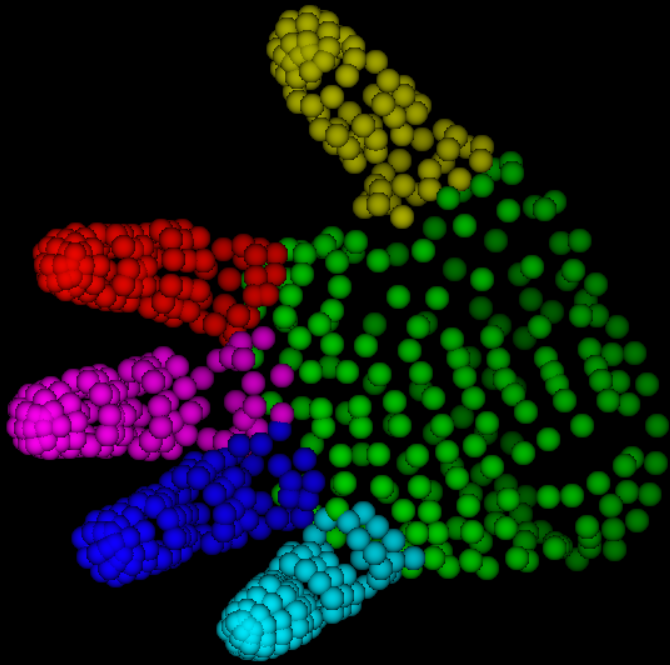}
    }
    ~~~~
    \subfigure[Depth map-based]{
     \includegraphics[width =  0.20\textwidth, height = 0.16\textwidth]{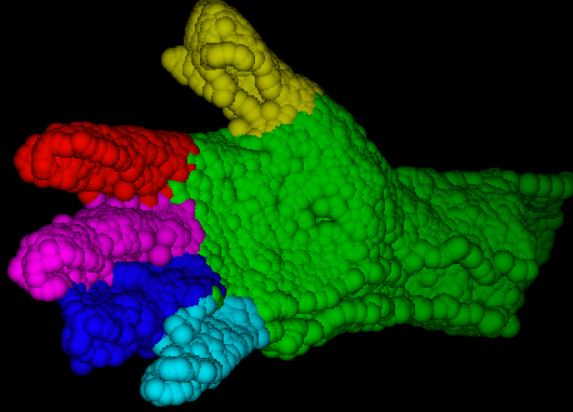}
    }~~~~
    \subfigure[Depth downsampled]{
     \includegraphics[width = 0.20\textwidth, height = 0.16\textwidth]{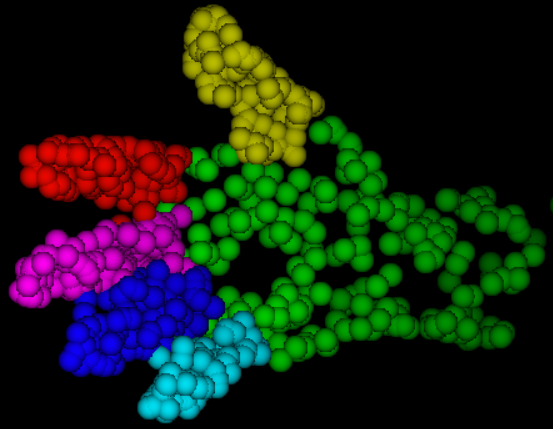}
    }
    \caption{Segmentation Transfer from the MANO vertices to real depth point clouds.
    }
    \label{fig:segmentation transfer}
    \vspace{-.6cm}
\end{figure}

\textbf{Local Component Assignment:} A local reconstruction requires assigning each template point to a fixed local component,~\ie which points in the ground truth should be used to evaluate each local reconstruction? For the MANO-generated point clouds, we manually split the 778 vertices into six semantic portions corresponding to the palm, thumb, and four fingers (see Fig.~\ref{fig:segmentation transfer}(a)). 
Real point clouds, however, are unordered. To assign points, we firstly estimate the 3D pose and the MANO parameters. We then apply a simple k-nearest neighbor classifier (k \!= 3) to match each point to a MANO vertex and give it the same component as the MANO vertices' partition. We can also re-sample as necessary (see Fig.~\ref{fig:segmentation transfer} (b) and (c)).
\subsection{MVHand Dataset}
Methods of generating point clouds based on MANO are straightforward approaches to add point cloud data to the current RGB datasets. However, they are not fully representational, as the MANO mesh is a smoothed approximation of a genuine hand's surface. Real-world point clouds from the multi-view depth maps are dense and noisy; variations between the two may be observed in Fig. \ref{fig:segmentation transfer}.  To verify that our framework also works on real-world point clouds, we need a multi-view RGB-D dataset,~\ie RGB images as input and multi-view depth for constructing ground truth point clouds. The only such dataset to date is the NYU Hand Pose Dataset~\cite{tompson2014real}.  However, this dataset does not provide camera extrinsics, so it is not possible to composite the views into complete point clouds. Furthermore, the recordings were done by Kinect V1 sensors and the depth maps are very noisy. 

\begin{figure}[t!]
\centering
\includegraphics[scale=0.38]{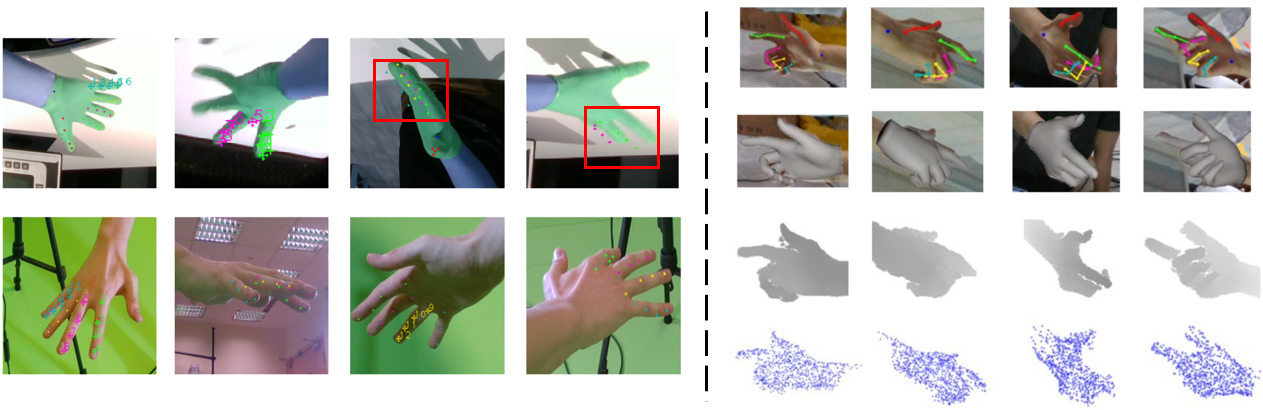}
\caption{Left: Different distances for annotation settings. Top row shows that recording with a larger distance between hand and camera. Bottom row shows that distance is smaller than 50 cm. Our manual annotations are based on the first column image and project into other three views. Right: Samples from our proposed MVHand dataset. We highlight the error region with red box.}
\label{fig:ourdataset}
\vspace{-0.2cm}
\end{figure}

To cover this gap, we record a new multi-view RGB-D dataset, which we call \href{https://github.com/ShichengChen/multiviewDataset}{MVHand}. Similar to the BigHand~\cite{yuan2017bighand2} and Ho3D~\cite{hampali2020honnotate} datasets, we record our dataset with Inter RealSense D415 cameras. We use four cameras, each at a range of approximately 50cm from the hand.  This is within the manufacturer's recommended range of 45cm to 2m, which is then specified to have a depth error of $<2\%$\footnote{https://www.intelrealsense.com/depth-camera-d415/}.  At this distance we observe that manually labelled 2D keypoints in one view (see Fig.~\ref{fig:ourdataset}) can accurately project onto the other views.

\begin{figure}[]
 \centering
 \includegraphics[scale=0.25]{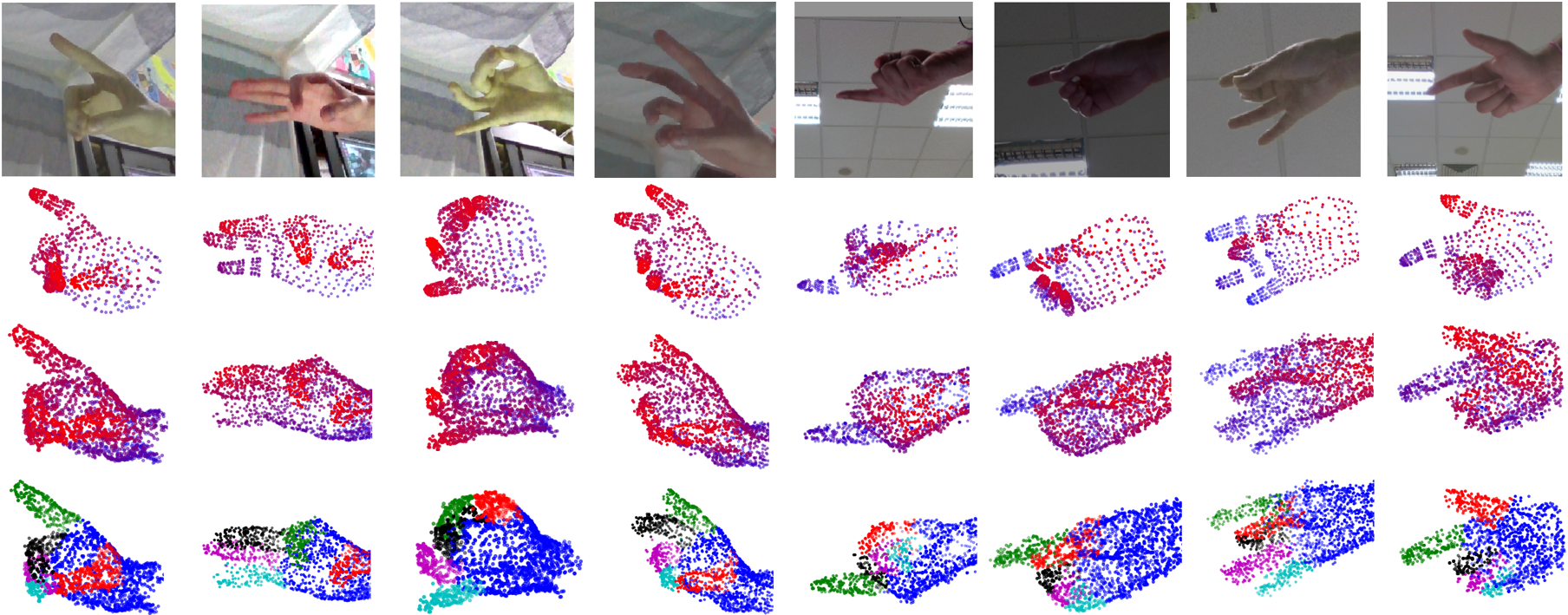}
 \caption{MVHand dataset point clouds visualization. These four rows show RGB images, mesh vertices, our complete point clouds, our point clouds segmentation by using K-Nearest Neighbors algorithm based on mesh vertices.}
\label{fig:quality}
\vspace{-0.35cm}
\end{figure}

To obtain the point clouds, we firstly segment the hand in the depth image via thresholding, and then project the four views into a complete hand. The final complete 3D point cloud ground truth is obtained by sampling from these depth maps. Specifically, we remove any isolated outlier points via filtering. Additionally, we downsample the points in overlapping areas from the different views to ensure that the points in the point cloud are evenly distributed.  Fig.~\ref{fig:quality} shows some sample point clouds from our dataset; our point clouds well-represent the original hand shape and are close to fitted MANO mesh vertices. The mean per-point Chamfer distance from point cloud to mesh vertex 7.88mm, with a standard deviation of 0.72 mm. We direct the reader to the Supplementary for further details. 

The 3D hand poses are annotated using the same 21-joint hand model as~\cite{zimmermann2017learning,sun2018integral}. We use a semi-automated method, combining human annotations (around 7\% of frames) with the self-supervision method of~\cite{wan2019dual} (see Supplementary 1.2 for details). The automatically labelled frames have an estimated average joint error of 4.69mm when evaluated against manually labelled samples, which is close to the manual label errors of Megahand~\cite{han2020megatrack}.

Table~\ref{tab:dataset} compares the statistics of MVHand with various hand benchmarks.  MVHand provides
RGB and depth maps from the four views, 3D joint positions, fitted MANO parameters, hand masks, and all camera intrinsics and extrinsics. Fig.~\ref{fig:ourdataset} shows sample frames; more visualizations can be found in the Supplementary.  

\begin{table}[]
\centering
\begin{tabular}{|c|c|c|c|c|c|}
\hline
& STB & RHD &FreiHand &YouTube3D &MVHand \\ \hline
Modality &real rgbd & syn. rgbd &real rgb &real rgb &real rgbd  \\ \hline
Resolution& $640\times480$  &$320\times 320$  &$240\times 240$   &mixed &$640\times480$ \\ \hline
Hands    &single &two  &single  &single  &single\\ \hline
Subjects  &1 &20   &32  &$-$ &4\\ \hline
Viewpoints & 2 & 1  &1  &1 & 4 \\ \hline
MANO params     &\xmark &\xmark &\cmark &\cmark  &\cmark\\ \hline
Frames   &36K &44K &36K &51K &83K\\ \hline
\end{tabular}
\caption{Comparison of proposed MVHand Dataset with other RGB benchmarks.}
\label{tab:dataset}
\vspace{-.3cm}
\end{table}   
\section{Experiments}
\textbf{Datasets and Metrics:} We test on four standard RGB-based hand post estimation benchmarks in addition to our own recorded MVhand dataset. \textbf{RHD}~\cite{zimmermann2017learning} is a synthetic dataset of rendered hands with 42k training images and 2.7k testing images. \textbf{STB}~\cite{zhang20163d} features videos of a single person's left hand in front of 6 real-world indoor backgrounds. We use the 15k/3k training/test split proposed in~\cite{zimmermann2017learning}. \textbf{FreiHAND}~\cite{zimmermann2019freihand} is a challenging multi-view RGB dataset of hand-object interactions. \textbf{YouTube3D} Hands-in-the-Wild~\cite{kulon2020weakly} features images curated from Youtube videos with a 47k/1.5k/1.5k image training/validation/test split. 

To evaluate 3D pose accuracy, we use area under the curve (AUC) on the percentage of the correct keypoint (PCK) score, where PCK is calculated using various error thresholds~\cite{zimmermann2017learning}. We also evaluate the mean 3D joint distance (mm) to ground truth according to mean-per-joint-position-error (MPJPE). To evaluate the reconstructed point clouds, we compute the mean Chamfer and Earth-Mover's distances as per Eq \ref{equ:cd} and \ref{equ:emd}.

\textbf{Implementation Details:} 
We optimize using ADAM to train the point cloud reconstruction firstly and then for 3D pose estimation. 
For RGB to point cloud encoder-decoder, we use an initial learning rate of 0.001, a weight decay of 1e-6 and a batch size of 32. Afterwards, we fine-tune the RGB encoder while learning the 3D pose decoder, using an initial {learning} rate of 0.0001 and weight decay of 1e-6.
\begin{table}
\begin{minipage}{0.42\linewidth}
\centering
\scalebox{0.93}{
\begin{tabular}{p{21mm}<{\centering}|p{9mm}<{\centering}|p{9mm}<{\centering}}
\hline
Method& RHD & STB \\ \hline
Zimm.'17 \cite{zimmermann2017learning} &$30.42$ &$8.68$     \\
Spurr'18 \cite{spurr2018cross}&$19.73$  &$8.56$       \\ 
Yang'19a \cite{yang2019disentangling}&$19.95$ &$8.56$   \\   
Bouk.'19 \cite{boukhayma20193d} &${16.78^*}$&${9.76}$\\ \hline
Yang'19b \cite{yang2019aligning}& $\textbf{13.14}$ &$7.05$  \\
Iqbal'18 \cite{iqbal2018hand} &$13.82$ &${8.01^*}$  \\ \hline
Ours(w/o rec.)   &$15.80$ &$7.72$ \\
Ours(full)&$13.38$  &$\textbf{6.71}$ \\ \hline
\end{tabular}}
\end{minipage}
\begin{minipage}{0.4\linewidth}
\scalebox{0.93}{
\begin{tabular}{p{21mm}<{\centering}|p{12mm}<{\centering}|p{15mm}<{\centering}|p{12mm}<{\centering}}
\hline
Method  &FreiHand  &YouTube3D &MVHand\\ \hline
Zimm.'19 \cite{zimmermann2019freihand}  &$11.0$ &$-$ &$-$  \\
Bouk.'19 \cite{boukhayma20193d} &${23.43^*}$ &${19.24^*}$  \\
 Chen'21 \cite{chen2021model} &$11.8$ &$-$ &$-$ \\
Choi'20 \cite{choi2020pose2mesh}  &\textbf{7.6} &$-$ &$-$ \\ \hline
Yang'19b \cite{yang2019aligning} &${12.35^*}$ &${18.76^*}$ &${15.12^*}$   \\
Iqbal'18 \cite{iqbal2018hand} &${13.52^*}$ &${19.32^*}$ &${15.27^*}$ \\ \hline
Ours(w/o rec.)    &$13.90$  &22.50 &17.80\\
Ours(full) &$9.60$ &$\textbf{18.50}$ &$\textbf{14.50}$\\ \hline
\end{tabular}}
\end{minipage}
\caption{Comparison of MPJPE (mm) with SOTA. "w/o rec." means without using our point cloud reconstruction pipeline, otherwise, "full". The best score is marked in \textbf{bold}. * indicates results based on released source code of~\cite{boukhayma20193d} and~\cite{yang2019aligning}, and our re-implementation of~\cite{iqbal2018hand}.}
\label{tab:pose estimation}
\vspace{-0.3cm}
\end{table}

\begin{figure}[htb]
	\centering
	\includegraphics[width=0.95\linewidth]{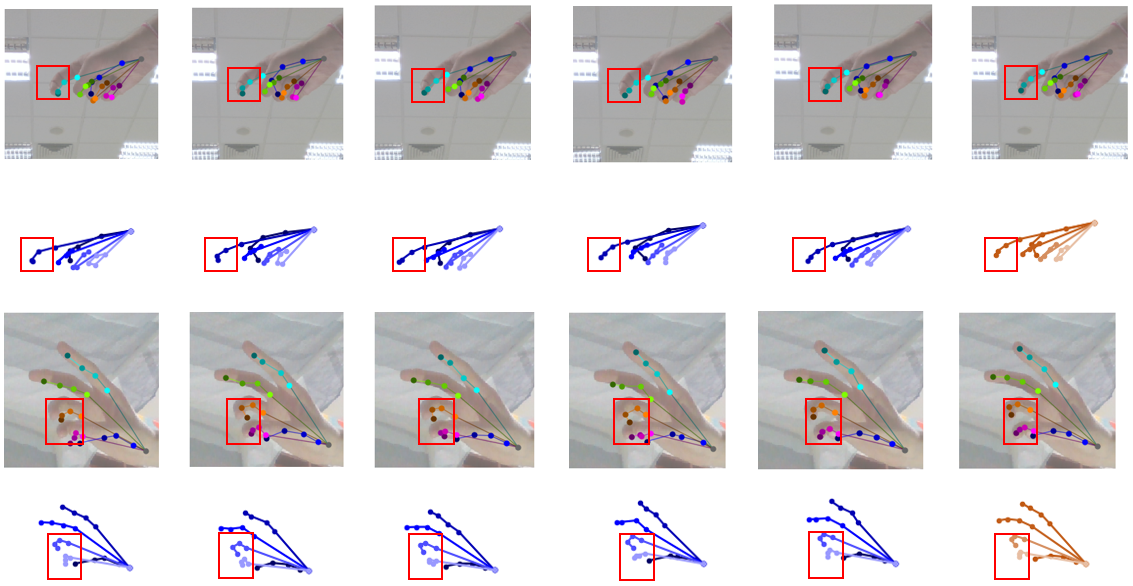}
	\caption{2D and 3D pose visualization. Left to right column:  Bouk.'19~\cite{boukhayma20193d},  Iqbal'18~\cite{iqbal2018hand}, Yang'19b~\cite{yang2019aligning}, Ours(w/o rec.), Ours(full), ground truth. We highlight the differences among predictions and the ground-truth poses with red boxes.}
	\label{fig:pose error1}
	\vspace{-0.6cm}
\end{figure}

\subsection{Comparison of Pose Estimates}
\begin{figure}[]
\centering
\subfigure[RHD AUC]{
\includegraphics[scale=0.4]{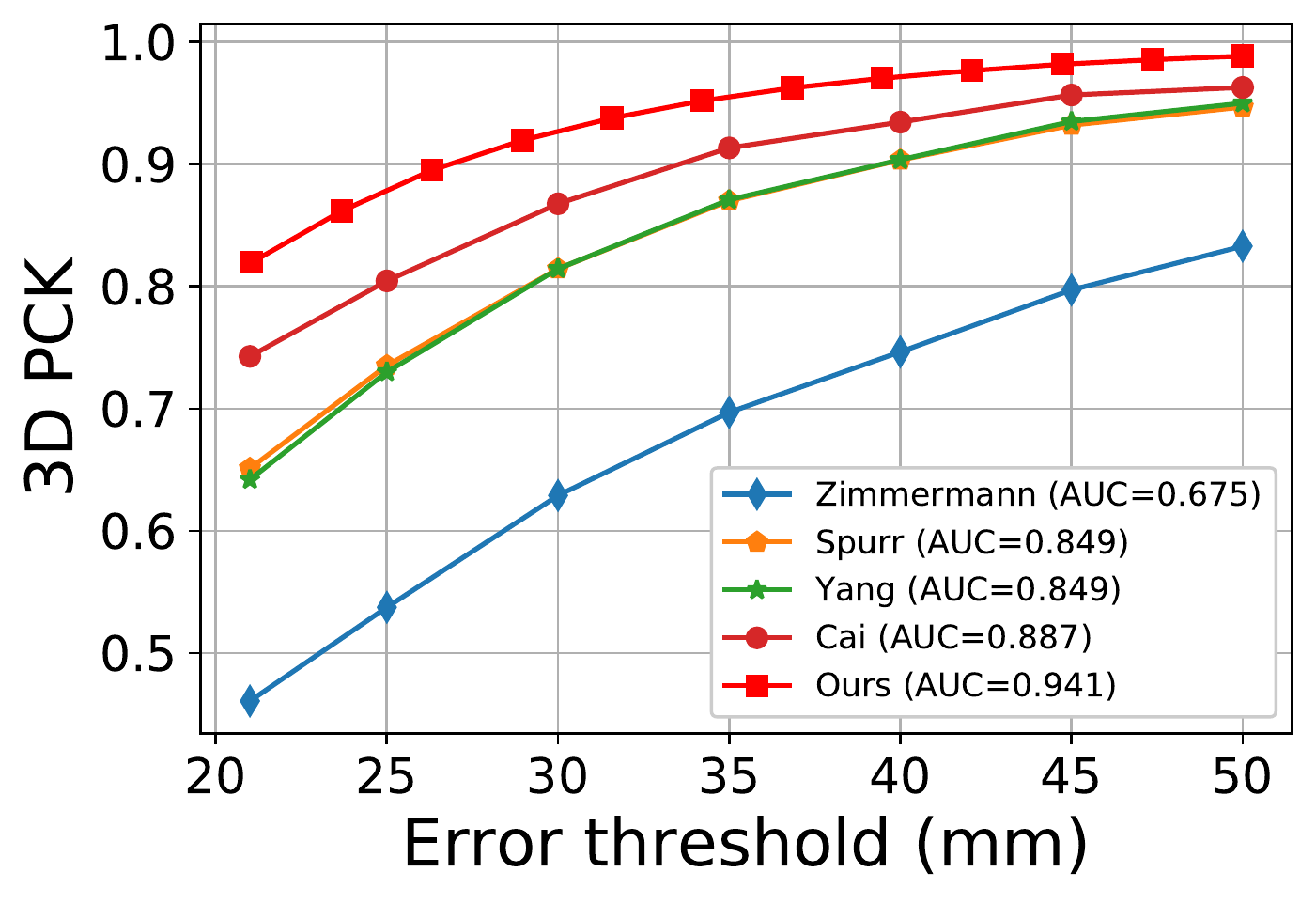}}
\subfigure[STB AUC]{
\includegraphics[scale=0.4]{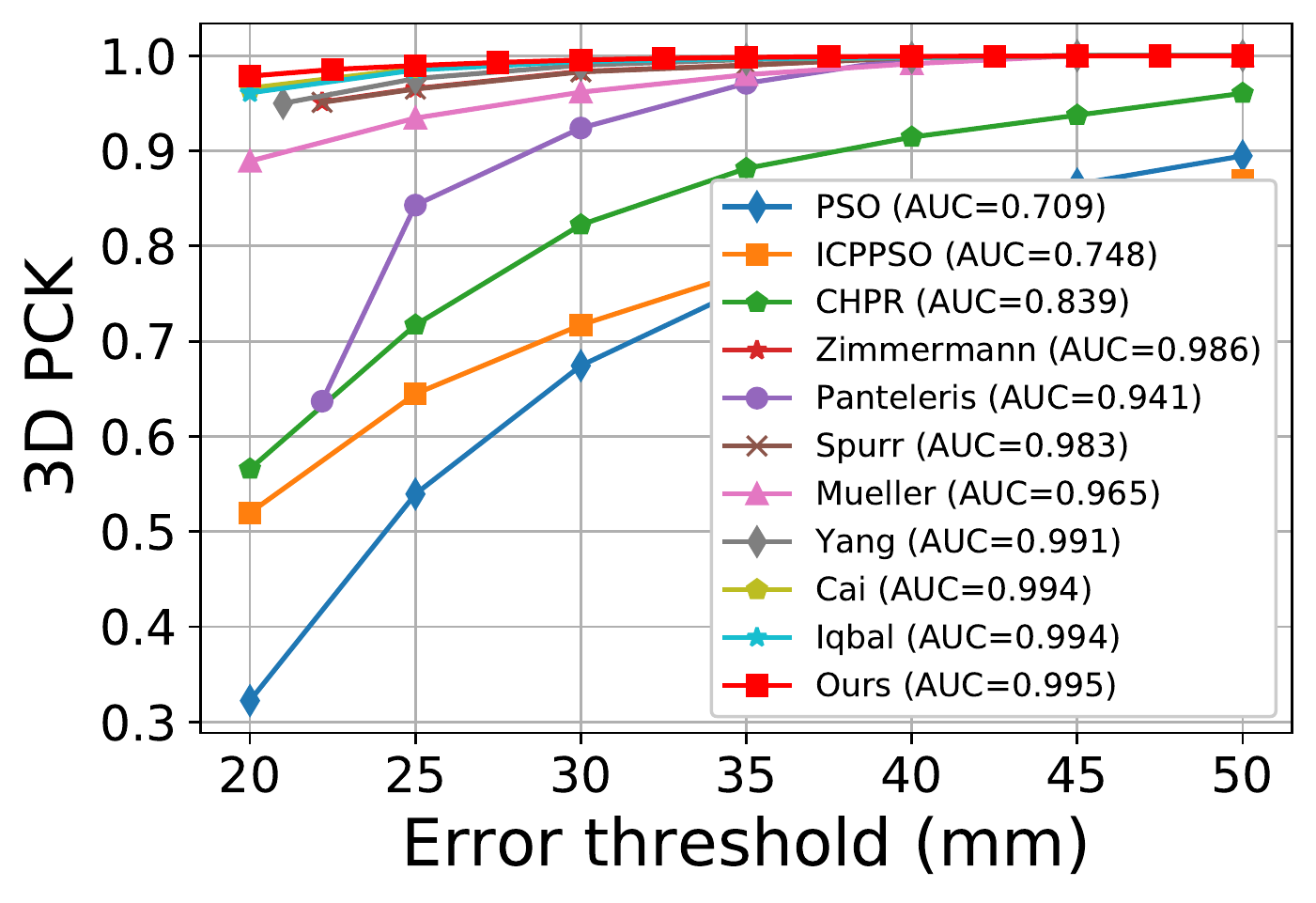}
}
\caption{Comparisons with state-of-the-art methods on RHD \cite{zimmermann2017learning} and STB \cite{zhang20163d} .}
\label{fig:RHD_AUC}
\vspace{-0.15cm}
\end{figure}

Table~\ref{tab:pose estimation} compares our 3D pose estimation accuracy. On RHD, our proposed framework surpasses most other approaches \cite{zimmermann2017learning, spurr2018cross, yang2019disentangling,iqbal2018hand}. Our MPJPE is comparable to Yang~\etal~\cite{yang2019aligning} despite their use of perspective correction and additional modalities like 2D heatmaps. On STB, we have the lowest MPJPE.  Figure~\ref{fig:RHD_AUC} (a) and (b) compare the 3D PCK curve on RHD and STB respectively. On both datasets, our proposed method obtains the highest AUC. There are very few published results on Freihand and YouTube3D. For FreiHand, our MPJPE (9.6mm) is lower than~\cite{zimmermann2019freihand, boukhayma20193d}. 
Choi'20 \cite{choi2020pose2mesh} reports 7.6mm, however, they use 2D poses from other pre-trained models as input. YouTube3D does not provide the hand scale, so we use 40mm as reference bone length \footnote{Reference bone length as defined by Freihand; 40mm comes from the STB dataset} to evaluate their test set. Table~\ref{tab:pose estimation} also compares pose estimation accuracy on our proposed MVHand dataset. As a baseline, we follow~\cite{iqbal2018hand} to directly regress the 3D hand pose with the 2.5D pose representation using a hold-one-subject out test split. Our full model's MPJPE, at 14.5mm, surpasses this baseline by 0.77mm.
More qualitative results are shown in Fig. \ref{fig:pose error1}.
\subsection{Ablation Studies}
\begin{table}
\centering
\begin{tabular}{c|c|c|c|c|c|c}
\hline
\multicolumn{4}{c|}{Chamfer Distance} & \multicolumn{3}{c}{Earth Mover's Distance}\\  \hline
           &2D grid & 3D hand  &local 3D   &2D grid & 3D hand  &local 3D \\ \hline
STB        &0.26    &0.25  &\textbf{0.24}   &1.75  & 1.71 &\textbf{1.68}  \\ \hline
RHD        &0.59       &0.55       &\textbf{0.45}   &2.01  &1.84  &\textbf{1.78}  \\ \hline
YouTube3D  &0.31    &0.30  &\textbf{0.27} &1.32 &1.24  &\textbf{0.99}   \\ \hline
MVHand        &1.20    &1.15   &\textbf{0.99}  &$-$ &$-$ &$-$\\ \hline
\end{tabular}
\caption{Mean CD and EMD per point; the best score is marked in \textbf{bold}.}
 \label{tab:cd_emd}
 \vspace{-0.35cm}
\end{table}
\textbf{Point Cloud Decoder:}
We remove the point cloud decoder and directly learn an image-to-pose encoder-decoder with the same architecture components as our current model.  Table \ref{tab:pose estimation} shows that the setting (w/o rec.) results in a higher error than the full model with the point-cloud decoder.  This is the case for all benchmark datasets and our MVHand; on average, the error is 20\% higher than the full model.  
These results verify that the point cloud reconstruction helps to learn a better latent representation for 3D pose estimation.
\begin{table}[htb]
\centering
\scalebox{0.85}{
 \begin{tabular}{c|cc|cc|cc|cc}
\hline
\multicolumn{5}{c|}{Chamfer Distance} & \multicolumn{4}{c}{Earth Mover's Distance}\\  \hline
          &\cite{boukhayma20193d} &Ours (3D) &\cite{yang2019aligning}(Sur.)    &Ours (Sur.)   &\cite{boukhayma20193d} &Ours (3D) &\cite{yang2019aligning}(Sur.)    &Ours (Sur.) \\ \hline
STB        &.367    &\textbf{.243}    &.146  &\textbf{.113}  &2.34 &\textbf{1.68} & \textbf{3.136} &5.294  \\ \hline
RHD        &.627    &\textbf{.450}    &\textbf{.195}  &.299 &\textbf{1.95} &1.78 &\textbf{4.434}  &5.134\\ \hline
\end{tabular}}
\caption{Mean CD and EMD per point; "Sur." indicates CD and EMD on a 2.5D space as per~\cite{iqbal2018hand}, since~\cite{yang2019aligning} estimates only the camera-facing surface of the hand. The best score is marked in \textbf{bold}. Surface result values are scaled by 1000.}
 \label{tab:MANO_recon}
 \vspace{-0.25cm}
\end{table}

\begin{figure}[htb]
\centering
\includegraphics[width=0.95\linewidth]{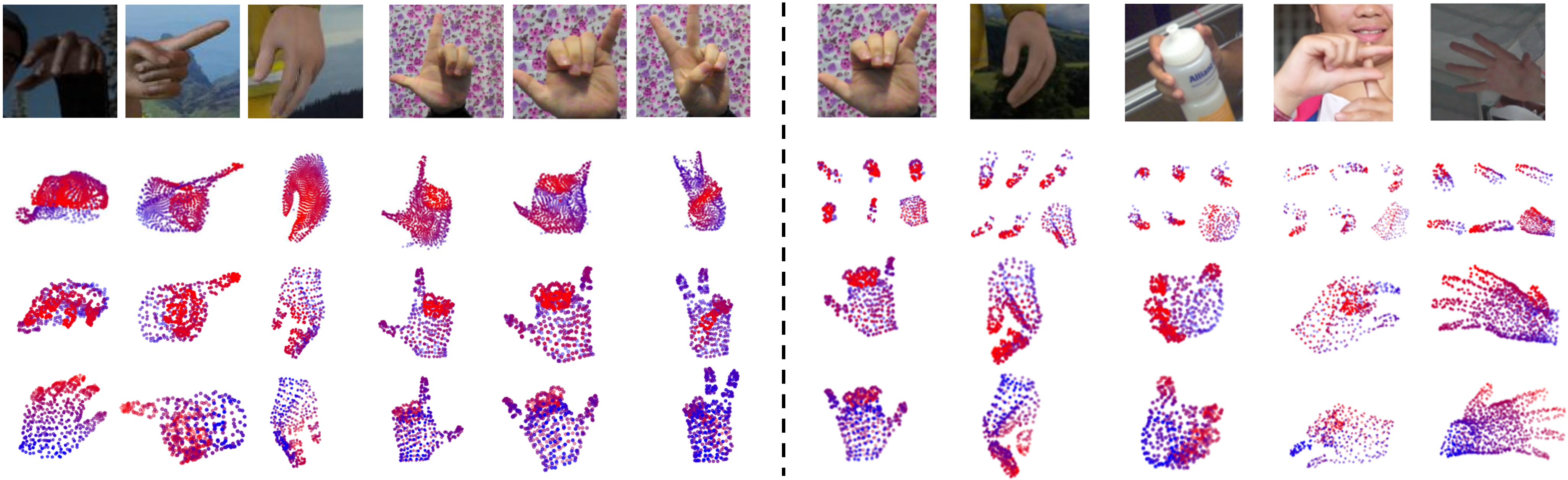}
\caption{Left: Reconstruction results compared with~\cite{yang2019aligning} for RHD and STB. Top to bottom: RGB images, incomplete reconstructions from~\cite{yang2019aligning}, our complete reconstruction from the same viewpoint and opposite viewpoint. Right:  Top to bottom: original RGB images from five datasets, the component-wise point cloud reconstructions from our method, aggregated into "Global" and its opposing view. Red points are close to the camera, and blue are far.}
\label{fig:compare1}
\vspace{-0.35cm}
\end{figure}

\textbf{Template Initialization:} We compare the point cloud reconstructions from the 2D grid, 3D hand, and local 3D hand initialization in Table~\ref{tab:cd_emd} using CD and EMD to evaluate; on both distances, a smaller value is better. We omit EMD on our MVHand dataset as the number of points (1038) is too large to compute within a feasible time. The results in Table~\ref{tab:cd_emd} support the strength of our local 3D hand initialization.

\subsection{Point Cloud Reconstructions} 
We visualize sample reconstructions in Figure~\ref{fig:compare1} for RHD and STB. Figure~\ref{fig:compare1} also compares to~\cite{yang2019aligning}'s method, which reconstructs the camera-facing surface point cloud. In Figure~\ref{fig:compare1}, we visualize samples from five datasets and observe that our reconstruction results are not only complete but also of higher quality, especially in distinguishing the individual fingers. Furthermore, we also visualize our component-wise point cloud reconstruction results and these results verify the effectiveness of our local and global framework in reconstructing high fidelity point clouds.  

As there are no other works that make complete point cloud reconstructions, we cannot make any direct quantitative comparisons.  Instead, we make two indirect comparisons in Table~\ref{tab:MANO_recon}.  First, we compare with~\cite{boukhayma20193d}, which directly regresses MANO parameters.  We project the estimated MANO parameters into a 3D point cloud in the same way as described in Sec.~\ref{vertice: gt}.  We find that our 3D results are in both CD and EMD for the STB dataset and better in CD for the RHD dataset. We also compare with~\cite{yang2019aligning}, though as their method only recovers surface poitn clouds, we project our complete point cloud into 2.5D space as per~\cite{iqbal2018hand}. Although our EMD results are worse than \cite{yang2019aligning}, we believe the projection process accumulates some errors.  When visualized, however, our point clouds are much cleaner and of higher quality than~\cite{yang2019aligning} (see Fig.~\ref{fig:compare1}).

\section{Conclusion}
This paper proposed a framework for reconstructing a complete 3D point cloud from RGB images.  In learning an RGB-point cloud encoder-decoder, we also learned a rich latent representation that can be decoded into an accurate 3D hand pose.
To improve the quality of the point clouds, we introduced two template initializations.  To verify our method on real-world hand point cloud data, we introduced MVHand, a new multi-view RGB-D dataset. Experimental results showed that our proposed method achieves comparable or better performance than existing 3D hand pose and shape estimation methods. In future work, we will explore the use of point clouds to resolve self-occlusions of the hand. 
\section{Acknowledgments}
This research is supported by Singapore Ministry of Education (MOE) Academic Research Fund Tier 1 T1251RES1819.
\bibliography{egbib}

\begin{thebibliography}{48}
\providecommand{\natexlab}[1]{#1}
\providecommand{\url}[1]{\texttt{#1}}
\expandafter\ifx\csname urlstyle\endcsname\relax
  \providecommand{\doi}[1]{doi: #1}\else
  \providecommand{\doi}{doi: \begingroup \urlstyle{rm}\Url}\fi

\bibitem[Alldieck et~al.(2019)Alldieck, Pons-Moll, Theobalt, and
  Magnor]{alldieck2019tex2shape}
Thiemo Alldieck, Gerard Pons-Moll, Christian Theobalt, and Marcus Magnor.
\newblock Tex2shape: Detailed full human body geometry from a single image.
\newblock In \emph{Proceedings of the IEEE/CVF International Conference on
  Computer Vision}, pages 2293--2303, 2019.

\bibitem[Boukhayma et~al.(2019)Boukhayma, Bem, and Torr]{boukhayma20193d}
Adnane Boukhayma, Rodrigo~de Bem, and Philip~HS Torr.
\newblock 3d hand shape and pose from images in the wild.
\newblock In \emph{Proceedings of the IEEE Conference on Computer Vision and
  Pattern Recognition}, pages 10843--10852, 2019.

\bibitem[Cai et~al.(2018)Cai, Ge, Cai, and Yuan]{cai2018weakly}
Yujun Cai, Liuhao Ge, Jianfei Cai, and Junsong Yuan.
\newblock Weakly-supervised 3d hand pose estimation from monocular rgb images.
\newblock In \emph{Proceedings of the European Conference on Computer Vision
  (ECCV)}, pages 666--682, 2018.

\bibitem[Choi et~al.(2020)Choi, Moon, and Lee]{choi2020pose2mesh}
Hongsuk Choi, Gyeongsik Moon, and Kyoung~Mu Lee.
\newblock Pose2mesh: Graph convolutional network for 3d human pose and mesh
  recovery from a 2d human pose.
\newblock In \emph{European Conference on Computer Vision}, pages 769--787.
  Springer, 2020.

\bibitem[Fan et~al.(2017)Fan, Su, and Guibas]{fan2017point}
Haoqiang Fan, Hao Su, and Leonidas~J Guibas.
\newblock A point set generation network for 3d object reconstruction from a
  single image.
\newblock In \emph{Proceedings of the IEEE conference on computer vision and
  pattern recognition}, pages 605--613, 2017.

\bibitem[Gadelha et~al.(2018)Gadelha, Wang, and
  Maji]{gadelha2018multiresolution}
Matheus Gadelha, Rui Wang, and Subhransu Maji.
\newblock Multiresolution tree networks for 3d point cloud processing.
\newblock In \emph{Proceedings of the European Conference on Computer Vision
  (ECCV)}, pages 103--118, 2018.

\bibitem[Ge et~al.(2016)Ge, Liang, Yuan, and Thalmann]{ge2016robust}
Liuhao Ge, Hui Liang, Junsong Yuan, and Daniel Thalmann.
\newblock Robust 3d hand pose estimation in single depth images: from
  single-view cnn to multi-view cnns.
\newblock In \emph{Proceedings of the IEEE conference on computer vision and
  pattern recognition}, pages 3593--3601, 2016.

\bibitem[Ge et~al.(2017)Ge, Liang, Yuan, and Thalmann]{ge20173d}
Liuhao Ge, Hui Liang, Junsong Yuan, and Daniel Thalmann.
\newblock 3d convolutional neural networks for efficient and robust hand pose
  estimation from single depth images.
\newblock In \emph{Proceedings of the IEEE Conference on Computer Vision and
  Pattern Recognition}, pages 1991--2000, 2017.

\bibitem[Ge et~al.(2018{\natexlab{a}})Ge, Cai, Weng, and Yuan]{ge2018hand}
Liuhao Ge, Yujun Cai, Junwu Weng, and Junsong Yuan.
\newblock Hand pointnet: 3d hand pose estimation using point sets.
\newblock In \emph{Proceedings of the IEEE Conference on Computer Vision and
  Pattern Recognition}, pages 8417--8426, 2018{\natexlab{a}}.

\bibitem[Ge et~al.(2018{\natexlab{b}})Ge, Ren, and Yuan]{ge2018point}
Liuhao Ge, Zhou Ren, and Junsong Yuan.
\newblock Point-to-point regression pointnet for 3d hand pose estimation.
\newblock In \emph{Proceedings of the European conference on computer vision
  (ECCV)}, pages 475--491, 2018{\natexlab{b}}.

\bibitem[Ge et~al.(2019)Ge, Ren, Li, Xue, Wang, Cai, and Yuan]{ge20193d}
Liuhao Ge, Zhou Ren, Yuncheng Li, Zehao Xue, Yingying Wang, Jianfei Cai, and
  Junsong Yuan.
\newblock 3d hand shape and pose estimation from a single rgb image.
\newblock In \emph{Proceedings of the IEEE conference on computer vision and
  pattern recognition}, pages 10833--10842, 2019.

\bibitem[Groueix et~al.()Groueix, Fisher, Kim, Russell, and
  Aubry]{groueix1802atlasnet}
T~Groueix, M~Fisher, VG~Kim, BC~Russell, and M~Aubry.
\newblock Atlasnet: a papier-m{\^a}ch{\'e} approach to learning 3d surface
  generation (2018).
\newblock \emph{arXiv preprint arXiv:1802.05384}, 11.

\bibitem[Hampali et~al.(2020)Hampali, Rad, Oberweger, and
  Lepetit]{hampali2020honnotate}
Shreyas Hampali, Mahdi Rad, Markus Oberweger, and Vincent Lepetit.
\newblock Honnotate: A method for 3d annotation of hand and object poses.
\newblock In \emph{Proceedings of the IEEE/CVF Conference on Computer Vision
  and Pattern Recognition}, pages 3196--3206, 2020.

\bibitem[Han et~al.(2020)Han, Liu, Cabezas, Twigg, Zhang, Petkau, Yu, Tai,
  Akbay, Wang, et~al.]{han2020megatrack}
Shangchen Han, Beibei Liu, Randi Cabezas, Christopher~D Twigg, Peizhao Zhang,
  Jeff Petkau, Tsz-Ho Yu, Chun-Jung Tai, Muzaffer Akbay, Zheng Wang, et~al.
\newblock Megatrack: monochrome egocentric articulated hand-tracking for
  virtual reality.
\newblock \emph{ACM Transactions on Graphics (TOG)}, 39\penalty0 (4):\penalty0
  87--1, 2020.

\bibitem[Iqbal et~al.(2018)Iqbal, Molchanov, Breuel Juergen~Gall, and
  Kautz]{iqbal2018hand}
Umar Iqbal, Pavlo Molchanov, Thomas Breuel Juergen~Gall, and Jan Kautz.
\newblock Hand pose estimation via latent 2.5 d heatmap regression.
\newblock In \emph{Proceedings of the European Conference on Computer Vision
  (ECCV)}, pages 118--134, 2018.

\bibitem[Joseph~Tan et~al.(2016)Joseph~Tan, Cashman, Taylor, Fitzgibbon,
  Tarlow, Khamis, Izadi, and Shotton]{joseph2016fits}
David Joseph~Tan, Thomas Cashman, Jonathan Taylor, Andrew Fitzgibbon, Daniel
  Tarlow, Sameh Khamis, Shahram Izadi, and Jamie Shotton.
\newblock Fits like a glove: Rapid and reliable hand shape personalization.
\newblock In \emph{Proceedings of the IEEE conference on computer vision and
  pattern recognition}, pages 5610--5619, 2016.

\bibitem[Khamis et~al.(2015)Khamis, Taylor, Shotton, Keskin, Izadi, and
  Fitzgibbon]{khamis2015learning}
Sameh Khamis, Jonathan Taylor, Jamie Shotton, Cem Keskin, Shahram Izadi, and
  Andrew Fitzgibbon.
\newblock Learning an efficient model of hand shape variation from depth
  images.
\newblock In \emph{Proceedings of the IEEE conference on computer vision and
  pattern recognition}, pages 2540--2548, 2015.

\bibitem[Klokov and Lempitsky(2017)]{klokov2017escape}
Roman Klokov and Victor Lempitsky.
\newblock Escape from cells: Deep kd-networks for the recognition of 3d point
  cloud models.
\newblock In \emph{Proceedings of the IEEE International Conference on Computer
  Vision}, pages 863--872, 2017.

\bibitem[Kulon et~al.(2020)Kulon, Guler, Kokkinos, Bronstein, and
  Zafeiriou]{kulon2020weakly}
Dominik Kulon, Riza~Alp Guler, Iasonas Kokkinos, Michael~M Bronstein, and
  Stefanos Zafeiriou.
\newblock Weakly-supervised mesh-convolutional hand reconstruction in the wild.
\newblock In \emph{Proceedings of the IEEE/CVF Conference on Computer Vision
  and Pattern Recognition}, pages 4990--5000, 2020.

\bibitem[Lang et~al.(2020)Lang, Manor, and Avidan]{lang2020samplenet}
Itai Lang, Asaf Manor, and Shai Avidan.
\newblock Samplenet: differentiable point cloud sampling.
\newblock In \emph{Proceedings of the IEEE/CVF Conference on Computer Vision
  and Pattern Recognition}, pages 7578--7588, 2020.

\bibitem[Li and Lee(2019)]{li2019point}
Shile Li and Dongheui Lee.
\newblock Point-to-pose voting based hand pose estimation using residual
  permutation equivariant layer.
\newblock In \emph{Proceedings of the IEEE Conference on Computer Vision and
  Pattern Recognition}, pages 11927--11936, 2019.

\bibitem[Li et~al.(2018)Li, Bu, Sun, Wu, Di, and Chen]{li2018pointcnn}
Yangyan Li, Rui Bu, Mingchao Sun, Wei Wu, Xinhan Di, and Baoquan Chen.
\newblock Pointcnn: Convolution on x-transformed points.
\newblock In \emph{Advances in neural information processing systems}, pages
  820--830, 2018.

\bibitem[Malik et~al.(2020)Malik, Abdelaziz, Elhayek, Shimada, Ali, Golyanik,
  Theobalt, and Stricker]{malik2020handvoxnet}
Jameel Malik, Ibrahim Abdelaziz, Ahmed Elhayek, Soshi Shimada, Sk~Aziz Ali,
  Vladislav Golyanik, Christian Theobalt, and Didier Stricker.
\newblock Handvoxnet: Deep voxel-based network for 3d hand shape and pose
  estimation from a single depth map.
\newblock In \emph{Proceedings of the IEEE/CVF Conference on Computer Vision
  and Pattern Recognition}, pages 7113--7122, 2020.

\bibitem[Mandikal et~al.(2018)Mandikal, KL, and Venkatesh~Babu]{mandikal20183d}
Priyanka Mandikal, Navaneet KL, and R~Venkatesh~Babu.
\newblock 3d-psrnet: Part segmented 3d point cloud reconstruction from a single
  image.
\newblock In \emph{Proceedings of the European Conference on Computer Vision
  (ECCV) Workshops}, pages 0--0, 2018.

\bibitem[Moon et~al.(2018)Moon, Yong~Chang, and Mu~Lee]{moon2018v2v}
Gyeongsik Moon, Ju~Yong~Chang, and Kyoung Mu~Lee.
\newblock V2v-posenet: Voxel-to-voxel prediction network for accurate 3d hand
  and human pose estimation from a single depth map.
\newblock In \emph{Proceedings of the IEEE conference on computer vision and
  pattern Recognition}, pages 5079--5088, 2018.

\bibitem[Mueller et~al.(2019)Mueller, Davis, Bernard, Sotnychenko, Verschoor,
  Otaduy, Casas, and Theobalt]{mueller2019real}
Franziska Mueller, Micah Davis, Florian Bernard, Oleksandr Sotnychenko, Mickeal
  Verschoor, Miguel~A Otaduy, Dan Casas, and Christian Theobalt.
\newblock Real-time pose and shape reconstruction of two interacting hands with
  a single depth camera.
\newblock \emph{ACM Transactions on Graphics (TOG)}, 38\penalty0 (4):\penalty0
  1--13, 2019.

\bibitem[Newell et~al.(2016)Newell, Yang, and Deng]{newell2016stacked}
Alejandro Newell, Kaiyu Yang, and Jia Deng.
\newblock Stacked hourglass networks for human pose estimation.
\newblock In \emph{European conference on computer vision}, pages 483--499.
  Springer, 2016.

\bibitem[Qi et~al.(2017)Qi, Su, Mo, and Guibas]{qi2017pointnet}
Charles~R Qi, Hao Su, Kaichun Mo, and Leonidas~J Guibas.
\newblock Pointnet: Deep learning on point sets for 3d classification and
  segmentation.
\newblock In \emph{Proceedings of the IEEE conference on computer vision and
  pattern recognition}, pages 652--660, 2017.

\bibitem[Romero et~al.(2017)Romero, Tzionas, and Black]{romero2017embodied}
Javier Romero, Dimitrios Tzionas, and Michael~J Black.
\newblock Embodied hands: Modeling and capturing hands and bodies together.
\newblock \emph{ACM Transactions on Graphics (ToG)}, 36\penalty0 (6):\penalty0
  245, 2017.

\bibitem[Sinha et~al.(2016)Sinha, Choi, and Ramani]{sinha2016deephand}
Ayan Sinha, Chiho Choi, and Karthik Ramani.
\newblock Deephand: Robust hand pose estimation by completing a matrix imputed
  with deep features.
\newblock In \emph{Proceedings of the IEEE conference on computer vision and
  pattern recognition}, pages 4150--4158, 2016.

\bibitem[Spurr et~al.(2018)Spurr, Song, Park, and Hilliges]{spurr2018cross}
Adrian Spurr, Jie Song, Seonwook Park, and Otmar Hilliges.
\newblock Cross-modal deep variational hand pose estimation.
\newblock In \emph{Proceedings of the IEEE Conference on Computer Vision and
  Pattern Recognition}, pages 89--98, 2018.

\bibitem[Sun et~al.(2018)Sun, Xiao, Wei, Liang, and Wei]{sun2018integral}
Xiao Sun, Bin Xiao, Fangyin Wei, Shuang Liang, and Yichen Wei.
\newblock Integral human pose regression.
\newblock In \emph{Proceedings of the European Conference on Computer Vision
  (ECCV)}, pages 529--545, 2018.

\bibitem[Tompson et~al.(2014)Tompson, Stein, Lecun, and
  Perlin]{tompson2014real}
Jonathan Tompson, Murphy Stein, Yann Lecun, and Ken Perlin.
\newblock Real-time continuous pose recovery of human hands using convolutional
  networks.
\newblock \emph{ACM Transactions on Graphics (ToG)}, 33\penalty0 (5):\penalty0
  1--10, 2014.

\bibitem[Wan et~al.(2020)Wan, Probst, Van~Gool, and Yao]{wan2019dual}
Chengde Wan, Thomas Probst, Luc Van~Gool, and Angela Yao.
\newblock Dual grid net: hand mesh vertex regression from single depth maps.
\newblock In \emph{ECCV}, 2020.

\bibitem[Xiong et~al.(2019)Xiong, Zhang, Xiao, Cao, Yu, Zhou, and
  Yuan]{xiong2019a2j}
Fu~Xiong, Boshen Zhang, Yang Xiao, Zhiguo Cao, Taidong Yu, Joey~Tianyi Zhou,
  and Junsong Yuan.
\newblock A2j: Anchor-to-joint regression network for 3d articulated pose
  estimation from a single depth image.
\newblock In \emph{Proceedings of the IEEE International Conference on Computer
  Vision}, pages 793--802, 2019.

\bibitem[Yang and Yao(2019)]{yang2019disentangling}
Linlin Yang and Angela Yao.
\newblock Disentangling latent hands for image synthesis and pose estimation.
\newblock In \emph{Proceedings of the IEEE Conference on Computer Vision and
  Pattern Recognition}, pages 9877--9886, 2019.

\bibitem[Yang et~al.(2019)Yang, Li, Lee, and Yao]{yang2019aligning}
Linlin Yang, Shile Li, Dongheui Lee, and Angela Yao.
\newblock Aligning latent spaces for 3d hand pose estimation.
\newblock In \emph{Proceedings of the IEEE International Conference on Computer
  Vision}, pages 2335--2343, 2019.

\bibitem[Yang et~al.(2018)Yang, Feng, Shen, and Tian]{yang2018foldingnet}
Yaoqing Yang, Chen Feng, Yiru Shen, and Dong Tian.
\newblock Foldingnet: Point cloud auto-encoder via deep grid deformation.
\newblock In \emph{Proceedings of the IEEE Conference on Computer Vision and
  Pattern Recognition}, pages 206--215, 2018.

\bibitem[Yao et~al.(2019)Yao, Fang, Wu, Feng, and Li]{yao2019densebody}
Pengfei Yao, Zheng Fang, Fan Wu, Yao Feng, and Jiwei Li.
\newblock Densebody: Directly regressing dense 3d human pose and shape from a
  single color image.
\newblock \emph{arXiv preprint arXiv:1903.10153}, 2019.

\bibitem[Ye et~al.(2018)Ye, Li, Huang, Du, and Zhang]{ye20183d}
Xiaoqing Ye, Jiamao Li, Hexiao Huang, Liang Du, and Xiaolin Zhang.
\newblock 3d recurrent neural networks with context fusion for point cloud
  semantic segmentation.
\newblock In \emph{Proceedings of the European Conference on Computer Vision
  (ECCV)}, pages 403--417, 2018.

\bibitem[Yuan et~al.(2017)Yuan, Ye, Stenger, Jain, and Kim]{yuan2017bighand2}
Shanxin Yuan, Qi~Ye, Bjorn Stenger, Siddhant Jain, and Tae-Kyun Kim.
\newblock Bighand2. 2m benchmark: Hand pose dataset and state of the art
  analysis.
\newblock In \emph{Proceedings of the IEEE Conference on Computer Vision and
  Pattern Recognition}, pages 4866--4874, 2017.

\bibitem[Yujin et~al.(2021)Yujin, Tu, Kang, Bao, Zhang, Zhe, Chen, and
  Yuan]{chen2021model}
Chen Yujin, Zhigang Tu, Di~Kang, Linchao Bao, Ying Zhang, Xuefei Zhe, Ruizhi
  Chen, and Junsong Yuan.
\newblock Model-based 3d hand reconstruction via self-supervised learning.
\newblock In \emph{Proceedings of the IEEE/CVF Conference on Computer Vision
  and Pattern Recognition}, pages 10451--10460, 2021.

\bibitem[Zeng and Gevers(2018)]{zeng20183dcontextnet}
Wei Zeng and Theo Gevers.
\newblock 3dcontextnet: Kd tree guided hierarchical learning of point clouds
  using local and global contextual cues.
\newblock In \emph{Proceedings of the European Conference on Computer Vision
  (ECCV)}, pages 0--0, 2018.

\bibitem[Zhang et~al.(2016)Zhang, Jiao, Chen, Qu, Xu, and Yang]{zhang20163d}
Jiawei Zhang, Jianbo Jiao, Mingliang Chen, Liangqiong Qu, Xiaobin Xu, and
  Qingxiong Yang.
\newblock 3d hand pose tracking and estimation using stereo matching.
\newblock \emph{arXiv preprint arXiv:1610.07214}, 2016.

\bibitem[Zhang et~al.(2020)Zhang, Huang, and Wang]{zhang2020object}
Tianshu Zhang, Buzhen Huang, and Yangang Wang.
\newblock Object-occluded human shape and pose estimation from a single color
  image.
\newblock In \emph{Proceedings of the IEEE/CVF Conference on Computer Vision
  and Pattern Recognition}, pages 7376--7385, 2020.

\bibitem[Zhang et~al.(2019)Zhang, Li, Mo, Zhang, and Zheng]{zhang2019end}
Xiong Zhang, Qiang Li, Hong Mo, Wenbo Zhang, and Wen Zheng.
\newblock End-to-end hand mesh recovery from a monocular rgb image.
\newblock In \emph{Proceedings of the IEEE International Conference on Computer
  Vision}, pages 2354--2364, 2019.

\bibitem[Zimmermann and Brox(2017)]{zimmermann2017learning}
Christian Zimmermann and Thomas Brox.
\newblock Learning to estimate 3d hand pose from single rgb images.
\newblock In \emph{Proceedings of the IEEE international conference on computer
  vision}, pages 4903--4911, 2017.

\bibitem[Zimmermann et~al.(2019)Zimmermann, Ceylan, Yang, Russell, Argus, and
  Brox]{zimmermann2019freihand}
Christian Zimmermann, Duygu Ceylan, Jimei Yang, Bryan Russell, Max Argus, and
  Thomas Brox.
\newblock Freihand: A dataset for markerless capture of hand pose and shape
  from single rgb images.
\newblock In \emph{Proceedings of the IEEE International Conference on Computer
  Vision}, pages 813--822, 2019.

\end{thebibliography}
\end{document}